%% file: FindTrack.tex
\documentclass[10pt,twocolumn,letterpaper]{article}

\usepackage{iccv}              


\definecolor{iccvblue}{rgb}{0.21,0.49,0.74}
\usepackage[pagebackref,breaklinks,colorlinks,allcolors=iccvblue]{hyperref}

\usepackage{hyperref}
\hypersetup{
colorlinks = true, 
urlcolor = magenta, 
linkcolor = red, 
citecolor = blue}
\usepackage{multirow}
\usepackage{enumitem}
\usepackage{romannum}
\usepackage{booktabs}
\usepackage{amsmath}
\usepackage{cuted}
\usepackage{capt-of}

\usepackage{array}
\newcolumntype{P}[1]{>{\centering\arraybackslash}p{#1}}

\usepackage{xcolor,colortbl}
\definecolor{Gray}{gray}{0.9}
\usepackage{algorithm}
\usepackage{algpseudocode}

\def\paperID{3} 
\def\confName{ICCV}
\def\confYear{2025}

\title{Find First, Track Next: Decoupling Identification and Propagation\\in Referring Video Object Segmentation}

\author{Suhwan Cho$^{1,*}$\quad Seunghoon Lee$^{2,*}$\quad Minhyeok Lee$^2$\quad Jungho Lee$^2$\quad Sangyoun Lee$^2$\vspace{0.2cm}\\
$^1$~GenGenAI\quad $^2$~Yonsei University\vspace{0.1cm}\\
\fontsize{10.0}{10.0}\url{https://github.com/suhwan-cho/FindTrack}\\
}

\begin{document}
\maketitle
\def\thefootnote{*}\footnotetext{These authors contribute equally to this work.}

\begin{strip}
\vspace{-1.8cm}
\centering
\includegraphics[width=1.0\textwidth]{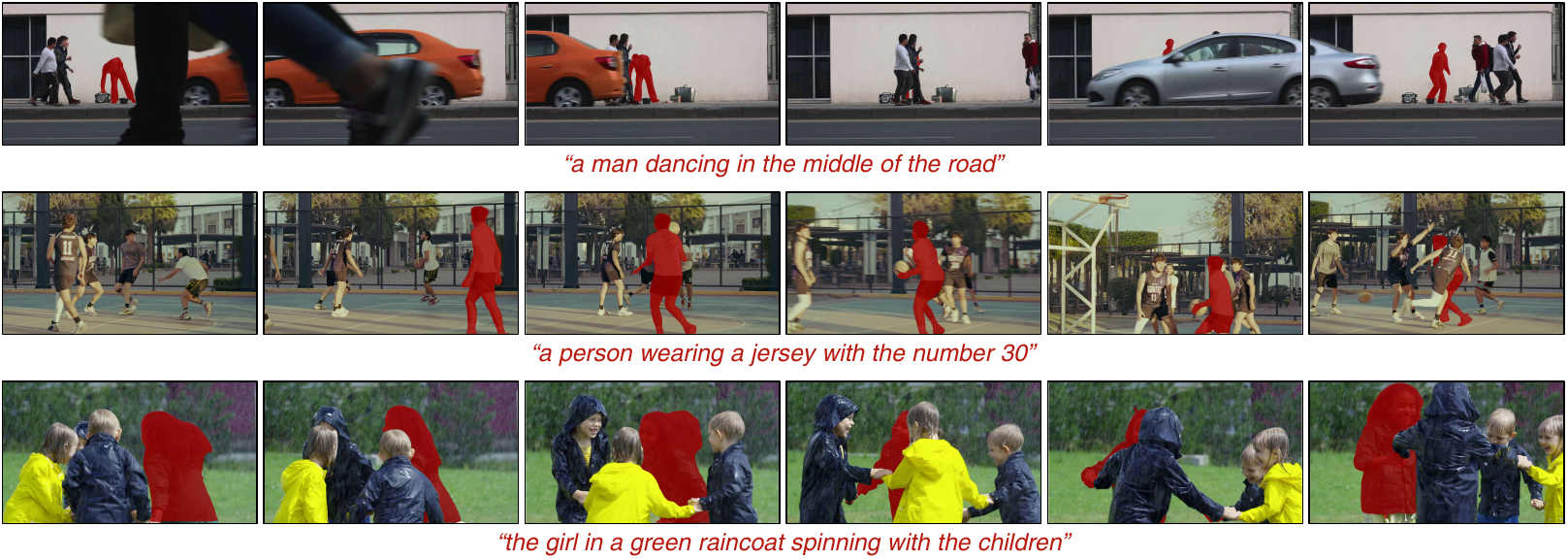}
\vspace{-6mm}
\captionof{figure}{Results on the real-world videos, illustrating the robust performance of FindTrack across challenging scenarios.\label{figure1}}
\end{strip}

\begin{abstract} 
Referring video object segmentation aims to segment and track a target object in a video using a natural language prompt. Existing methods typically fuse visual and textual features in a highly entangled manner, processing multi-modal information together to generate per-frame masks. However, this approach often struggles with ambiguous target identification, particularly in scenes with multiple similar objects, and fails to ensure consistent mask propagation across frames. To address these limitations, we introduce FindTrack, an efficient decoupled framework that separates target identification from mask propagation. FindTrack first adaptively selects a key frame by balancing segmentation confidence and vision-text alignment, establishing a robust reference for the target object. This reference is then utilized by a dedicated propagation module to track and segment the object across the entire video. By decoupling these processes, FindTrack effectively reduces ambiguities in target association and enhances segmentation consistency. FindTrack significantly outperforms all existing methods on public benchmarks, demonstrating its superiority.
\vspace{-0.5cm}
\end{abstract}

\begin{figure*}[t]
\centering
\includegraphics[width=1\linewidth]{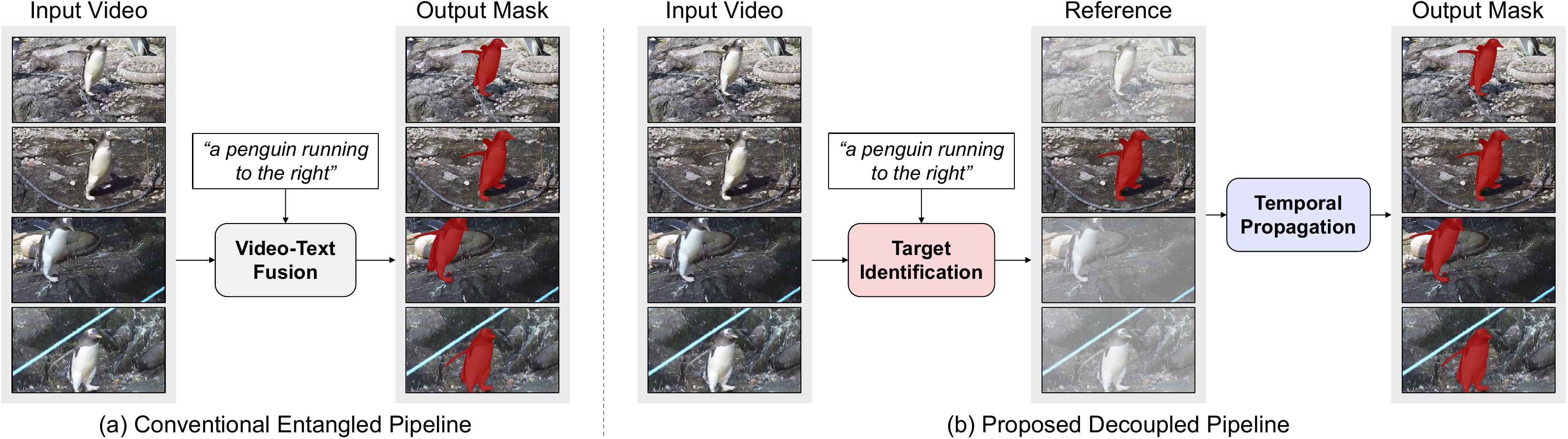}
\vspace{-6mm}
\caption{Visualized comparison of (a) a conventional entangled pipeline that fuses multi-modal cues in a single stage and (b) our proposed decoupled pipeline, which separates target object identification and temporal mask propagation into two distinct stages.}
\label{figure2}
\end{figure*}

\section{Introduction}
Video object segmentation (VOS) aims to segment objects in videos and encompasses several settings. Semi-supervised VOS segments objects with an initial frame mask as guidance~\cite{STM, KMN, BMVOS, TBD, XMem}, while unsupervised VOS identifies primary objects without explicit supervision~\cite{MATNet, FSNet, TMO, PMN, GSA-Net, DPA}. Weakly-supervised VOS relies on a bounding box instead of a mask in the first frame~\cite{SiamMask, LWL, QMRA}. In contrast, referring VOS segments an object based on a natural language prompt, requiring the model to understand both visual and textual information. This task has gained significant interest due to its applications in video editing, human-computer interaction, and multimedia analysis. However, it remains challenging due to linguistic ambiguities, occlusions, and rapid motion.

Existing referring VOS methods predominantly follow an entangled fusion strategy, where visual and textual features are jointly processed within a spatio-temporal framework~\cite{MTTR, ReferFormer, SgMg, LoSh}. Typically, visual features are extracted via deep video encoders, while textual representations are obtained from language encoders. These multi-modal features are fused through attention mechanisms to generate per-frame segmentation masks. While effective, this approach has fundamental limitations. First, it often struggles with ambiguous target identification, particularly in scenes with multiple similar objects, as the model lacks an explicit reference frame to associate the language prompt with the correct object. Second, it does not enforce consistent object segmentation across frames, leading to propagation errors, especially in challenging scenarios involving occlusions, rapid motion, or appearance variations. Without a stable reference frame, existing methods fail to ensure robust and coherent mask propagation.

To address these challenges, we introduce FindTrack, an efficient decoupled framework that explicitly separates object identification from mask propagation. Rather than jointly processing all frames with fused multi-modal features, FindTrack first selects a key frame adaptively by considering segmentation confidence and vision-text alignment, ensuring a reliable and unambiguous reference for the target object. This reference frame serves as a stable anchor before mask propagation begins. A dedicated propagation module then leverages this strong reference to track and segment the target object throughout the video. By decoupling these two processes, FindTrack mitigates ambiguities in object identification and enhances segmentation consistency, overcoming the limitations of entangled feature fusion in existing methods. A visual comparison between FindTrack and existing approaches is shown in Figure~\ref{figure2}.

This decoupled framework offers several key advantages. First, by establishing a strong reference before propagation, FindTrack significantly improves target localization, reducing errors introduced by multi-modal fusion across all frames. Second, the use of a dedicated tracking mechanism ensures robust object tracking even in challenging scenarios such as occlusions and fast motion. Third, by explicitly modeling the transition from object identification to mask propagation, FindTrack provides a more structured and interpretable approach to referring VOS, offering insights that can guide future research in this domain.

Our proposed FindTrack is evaluated on three widely used referring VOS benchmarks: Ref-YouTube-VOS~\cite{URVOS}, Ref-DAVIS17~\cite{RefDAVIS}, and MeViS~\cite{MeViS}. On all datasets, FindTrack achieves state-of-the-art performance, significantly outperforming existing methods. Qualitative results on real-world videos (Figure~\ref{figure1}) further underscore the practical effectiveness of the proposed method.

\begin{figure*}[t]
\centering
\includegraphics[width=1\linewidth]{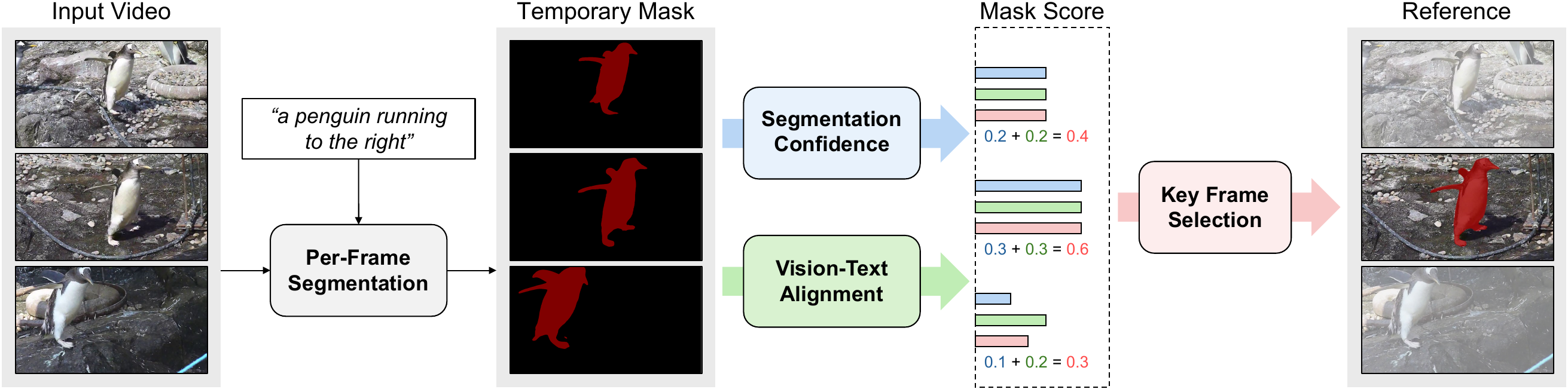}
\vspace{-6mm}
\caption{Visual representation of the target object identification process, the first step of FindTrack. Given an input video, candidate key frames are sampled to evaluate frame-level segmentation quality. The key frame is then selected based on segmentation confidence and vision-text alignment, serving as a reference for the next step.}
\label{figure3} 
\end{figure*}

Our main contributions are summarized as follows: 
\begin{itemize}[leftmargin=0.2in] 
\item We propose FindTrack, an efficient framework that explicitly decouples object identification from mask propagation, overcoming the limitations of entangled fusion.
\item We introduce an adaptive key frame selection strategy that leverages segmentation confidence and text alignment to establish a reliable anchor for propagation.
\item FindTrack achieves state-of-the-art performance on public benchmarks, outperforming existing methods.
\end{itemize}

\section{Related Work}

\noindent\textbf{Referring image segmentation.}
Referring image segmentation (RIS) focuses on segmenting the object in an image described by a natural language expression. Early methods emphasize the fusion of visual and linguistic cues. For instance, VLT~\cite{VLT, VLT2} introduces a vision-language transformer that generates object queries by independently extracting and fusing visual and textual features, enabling robust cross-modal interaction. CRIS~\cite{CRIS} leverages the strong image-text alignment of CLIP~\cite{CLIP} through pixel-level contrastive learning to distinguish subtle differences between the referred object and distractors. LAVT~\cite{LAVT} refines this fusion by integrating linguistic information at intermediate levels within the vision transformer. GRES~\cite{GRES} pushes RIS towards broader generalization and practical applicability. Recent advances, such as GSVA~\cite{GSVA}, exploit the generalization capabilities of large-scale multi-modal models to generate effective referring prompts, while EVF-SAM~\cite{EVF-SAM} employs an early fusion strategy to guide segmentation via elaborated prompts for SAM~\cite{SAM}.

\vspace{1mm}
\noindent\textbf{Referring video object segmentation.}
Referring VOS extends RIS to the video domain, requiring both accurate spatial segmentation and temporal consistency. Early methods focus on directly injecting text features into the visual stream. URVOS~\cite{URVOS} introduces a unified framework that combines an image-language fusion module with a memory mechanism to ensure consistent segmentation over time, whereas CMPC-V~\cite{CMPC} employs cross-modal attention across layers to establish fine-grained relationships between textual cues and visual features.

With the advent of DETR-based architectures~\cite{DETR}, query-based models have significantly advanced referring VOS. MTTR~\cite{MTTR} presents an end-to-end framework in which object-representative queries are generated and fused with multimodal features via transformers, with segmentation achieved via Hungarian matching. Building on this, ReferFormer~\cite{ReferFormer} leverages language as queries to interact with video features through cross-attention, further refined by a cross-modal FPN~\cite{FPN} decoder. SgMg~\cite{SgMg} mitigates feature drift by incorporating Fourier domain attention with a Gaussian kernel, maintaining consistency across frames, while VD-IT~\cite{VD-IT} exploits pre-trained text-to-video diffusion models~\cite{VDM, VDM2} within a query-based matching framework that combines CLIP-generated prompts with DETR.

Recent efforts aim to better integrate linguistic cues by disentangling static attributes from dynamic features. LoSh~\cite{LoSh} adopts a joint prediction strategy that categorizes language expressions into long and short components prior to fusion with visual features, and DsHmp~\cite{DsHmp} introduces a hierarchical perception module that decomposes language into static and motion parts while segmenting the target object into multiple regions for fine-grained alignment.

Despite these advancements, existing methods still face challenges. Entangled fusion strategies, which process visual and textual features jointly across all frames, can lead to ambiguous target identification in scenes with similar objects. Moreover, without a stable reference for segmentation, these approaches often suffer from propagation errors under occlusions or fast motion. Our work, FindTrack, addresses these limitations by explicitly decoupling object identification from mask propagation, thereby establishing a reliable reference and ensuring stable tracking of the target object across frames.

\section{Approach}

\subsection{Problem Formulation}
The objective of referring VOS is to segment the target object specified by a given language expression across all frames of a video. Formally, let the input video frames be denoted as $I := \{I^1, I^2, ..., I^T\}$, where $T$ is the total number of frames. The corresponding ground truth binary masks are represented as $M := \{M^1, M^2, ..., M^T\}$, and the language expression describing the target object is given by $L$. The goal is to predict a sequence of segmentation masks, $\hat{M} := \{\hat{M}^1, \hat{M}^2, ..., \hat{M}^{T}\}$, that accurately localizes the referred object in each frame based on $I$ and $L$.

\subsection{System Overview}
Existing methods fuse visual and textual features in a single-stage manner, as shown in Figure~\ref{figure2} (a), often leading to ambiguous target identification and inconsistent object tracking. To overcome these limitations, we propose a decoupled pipeline that explicitly separates target identification from temporal propagation, as illustrated in Figure~\ref{figure2} (b). Given video frames and a language expression, our approach first selects a key frame and predicts its segmentation mask. This mask is then bi-directionally propagated across the entire video, ensuring robust and consistent tracking of the target object specified by the text input.

\subsection{Target Identification}
The objective of target identification is to determine the target object specified by the language prompt within a video. This step focuses on localizing the most confident object in a single frame, rather than predicting the object mask across all frames, as the designated object will be tracked in the subsequent step. A visual representation of the target identification pipeline is shown in Figure~\ref{figure3}.

\vspace{1mm}
\noindent\textbf{Per-frame segmentation.}
To localize the target object with the highest confidence, we scan the video frames and predict the segmentation mask based on the language prompt. However, processing all frames incurs significant computational costs. To address this, we first sample candidate key frames and perform per-frame segmentation on them. The candidate key frames are defined as $C:=\{j\}_{i=1}^{N}$ where
\begin{align}
&j = \lfloor\frac{(i - 1) * (T - 1)}{N - 1}\rfloor + 1~,
\label{eq1}
\end{align}
and $N$ denotes the total number of sampled frames. For the uniformly sampled candidate key frames $C$, we apply a per-frame segmentation model, EVF-SAM~\cite{EVF-SAM}, resulting in a temporary mask set $\tilde{M}$ and the corresponding segmentation confidence $\pi$:
\begin{align}
&\tilde{M}, \pi := \{\Psi(I^j, L)\}^N_{i=1}~,
\label{eq2}
\end{align}
where $\Psi$ represents the per-frame referring segmentation model. The frame indices $j$ are determined as per Eqn.\ref{eq1}.

\begin{figure}[t]
\centering
\includegraphics[width=0.9\linewidth]{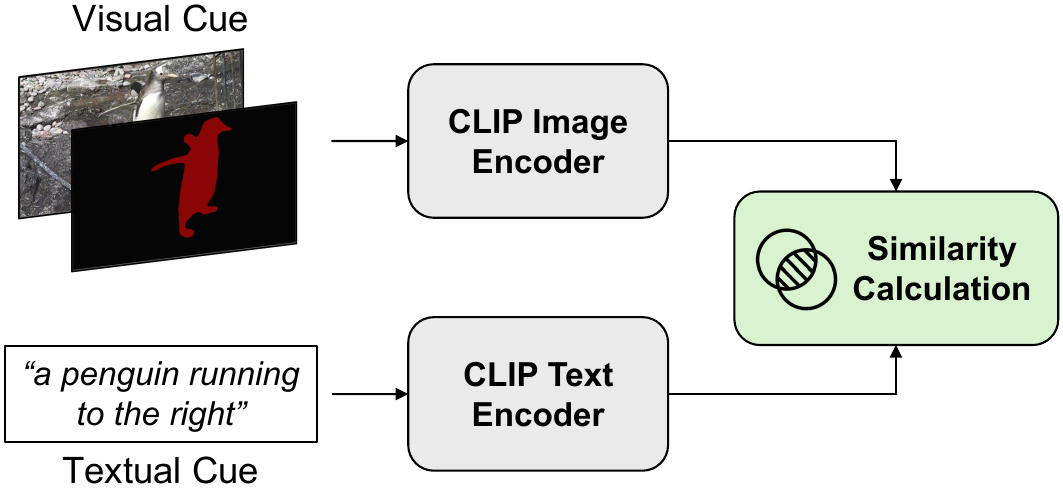}
\vspace{-1mm}
\caption{Visualized pipeline of the vision-text alignment stage. The similarity between the pixels within the predicted mask and the given text is evaluated to ensure contextual alignment.}
\label{figure4}
\end{figure}

\vspace{1mm}
\noindent\textbf{Key frame selection.}
From the temporary mask set $\tilde{M}$, we select the most reliable prediction as the key frame, which serves as a strong reference for target object tracking. The quality of each segmentation mask is assessed based on two criteria: 1) the segmentation confidence of the predicted mask and 2) the vision-text alignment between the visual content within the mask and the input language expression.

Segmentation confidence $\pi$ represents the predicted intersection over union (IoU) score of the generated mask. This confidence is learned alongside mask prediction during the training of the referring segmentation model. A high segmentation confidence suggests that the mask is likely to have a high IoU with the ground truth, thereby increasing the likelihood that the selected key frame will serve as an effective reference during temporal propagation. This confidence is directly derived from the segmentation decoder, as indicated in Eqn.~\ref{eq2}, which we employ as our per-frame referring segmentation model.

However, in some cases, even a high-confidence mask may fail to accurately localize the object with respect to the language prompt. To address this, we employ Alpha-CLIP~\cite{AlphaCLIP} to estimate the vision-text alignment score, ensuring that the selected key frame is not only reliable in segmentation but also well-aligned with the text prompt:
\begin{align}
&\rho = \mathcal{N}(\text{CLIP}_{img} (I \oplus \tilde{M})) \cdot \mathcal{N}(\text{CLIP}_{text})~,
\label{eq3}
\end{align}
where $\text{CLIP}_{img}$ and $\text{CLIP}_{text}$ denote the image and text encoders of Alpha-CLIP, respectively, and $\mathcal{N}$ represents L2 normalization along the channel dimension. By computing the cosine similarity between the visual and textual embeddings, we assess how well the visual content within the predicted mask aligns with the textual description. This allows us to verify whether the target object is correctly localized. This process is also visualized in Figure~\ref{figure4}. Finally, the overall mask score for each frame $\sigma$ is computed as
\begin{align}
&\sigma = \mathrm{w}_1 \pi + \mathrm{w}_2 \rho~,
\label{eq4}
\end{align}
where $\mathrm{w}_1$ and $\mathrm{w}_2$ denotes the weighting factor for the segmentation confidence score and vision-text alignment score, respectively. The final mask score serves as the criterion for key frame selection, defined as
\begin{align}
&k = \{\underset{j}{\arg\max}~\sigma^j\}^N_{i=1}~,
\label{eq5}
\end{align}
where $k$ represents the key frame used as a reference for temporal propagation.

\begin{figure}[t]
\centering
\includegraphics[width=0.85\linewidth]{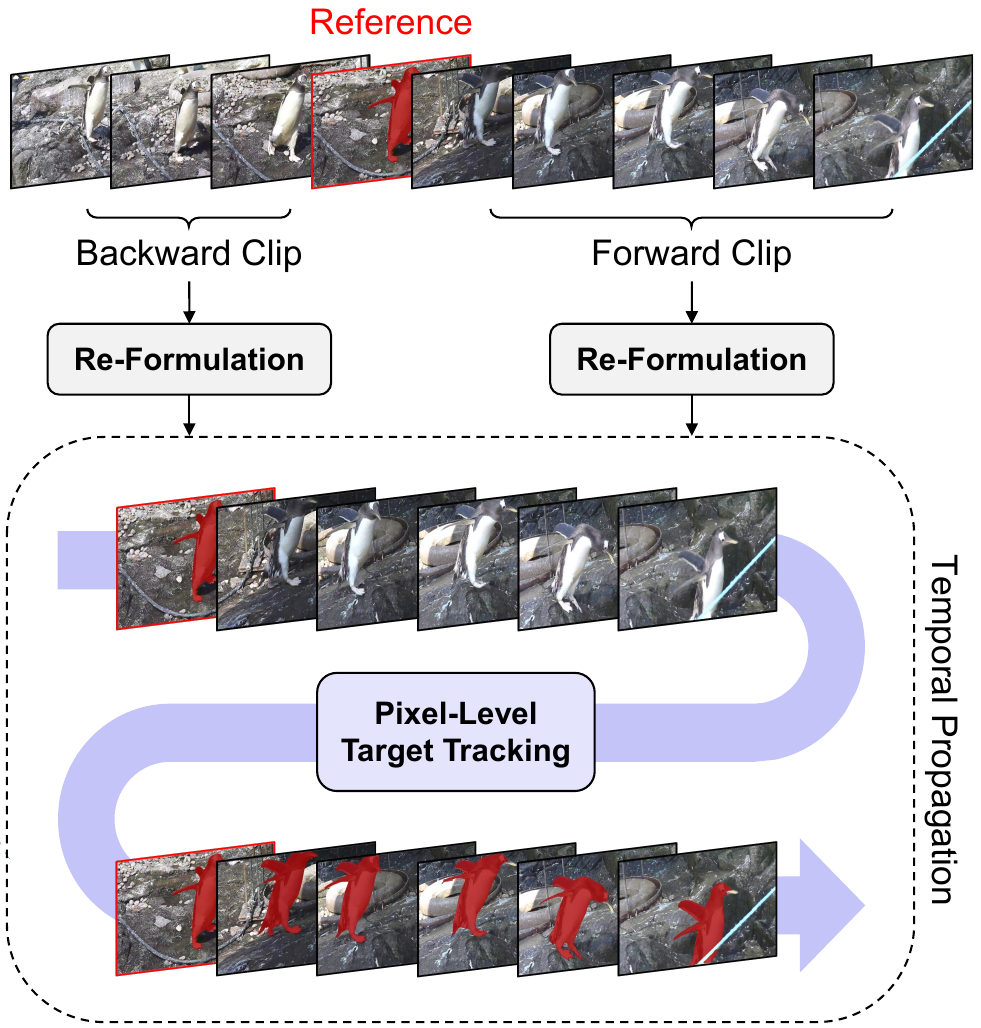}
\vspace{-1mm}
\caption{Visualized pipeline of temporal mask propagation. The video is split into two clips at the key frame, with each clip processed by the pixel-level tracking network.}
\label{figure5}
\end{figure}

\subsection{Temporal Propagation}
In the temporal mask propagation stage, the key frame selected in the previous step serves as a reference for propagating the target object's mask across subsequent video frames. This propagation exploits the temporal coherence of the object and the visual context in adjacent frames to ensure accurate localization throughout the sequence. The core idea is to transfer the target object's mask from the key frame while preserving consistency. To achieve this, we utilize the architecture of Cutie~\cite{Cutie}. The overall pipeline for temporal propagation is visualized in Figure~\ref{figure5}.

\vspace{1mm}
\noindent\textbf{Sequence re-formulation.}
Unlike existing approaches that directly detect the target object without an explicit visual reference, we track the object using a strong reference, specifically the predicted mask at the key frame. To exploit the visual prior of the target object, we first split the video into two clips:
\begin{align}
&I_{fw} := \{I^k, I^{k+1}, ..., I^T\}\\
&I_{bw} := \{I^k, I^{k-1}, ..., I^1\}~,
\label{eq6}
\end{align}
where $I_{fw}$ and $I_{bw}$ denote the forward and backward video clips, respectively. The first frame in each clip is initialized with the segmentation mask, providing strong guidance for tracking the target object.

\vspace{1mm}
\noindent\textbf{Pixel-level tracking.}
To achieve precise mask propagation, we employ a memory-based pixel tracking network~\cite{STM, STCN, XMem, Cutie}, as detailed in Algorithm~\ref{algorithm1}. This architecture, commonly used in semi-supervised VOS, maintains a dynamic memory bank that continuously stores and retrieves object features, ensuring accurate and temporally consistent segmentation throughout the video. By leveraging memory-based tracking, our approach effectively handles challenges such as object deformation and occlusions, which often degrade segmentation performance in dynamic scenes.

The process begins by initializing the memory with the key frame image $I^k$ and its predicted mask $\tilde{M}^k$. The memory module captures rich visual embeddings of the target object, which are then retrieved to guide mask prediction in subsequent frames. Propagation follows a bi-directional strategy, where the forward clip $I_{fw}$ and backward clip $I_{bw}$ are processed independently. At each step, the model predicts the segmentation mask for the current frame based on stored memory representations and updates the memory with the newly processed frame and its mask. This iterative update mechanism enhances robustness to appearance changes and complex motion patterns, ensuring stable mask propagation even in challenging scenarios with background clutter or fast-moving objects.

By structuring mask propagation within a memory-based tracking framework, our approach reliably transfers the key frame mask across the entire video sequence. The model continuously refines segmentation accuracy by leveraging both spatial and temporal cues accumulated over time. As a result, we obtain the final segmentation mask prediction $\hat{M}$, providing temporally coherent and spatially precise object tracking across all frames.

\begin{algorithm}[t!]
\small
\caption{Pixel-Level Tracking}
\begin{algorithmic}[1]
\State \textbf{Input:} $I$, $\tilde{M}^k$
\State \textbf{Output:} $\hat{M}$
\State Initialize $\hat{M}^k$ with $\tilde{M}^k$
\State Initialize memory with ($I^k, \tilde{M}^k$)
\For{each frame $t$ in \{k+1, k+2, ...,T\}}
\State Predict $\hat{M}^t$ using $I^t$ and memory
\State Update memory with ($I^t, \hat{M}^t$)
\EndFor
\State Initialize memory with ($I^k, \tilde{M}^k$)
\For{each frame $t$ in \{k-1, k-2, ...,1\}}
\State Predict $\hat{M}^t$ using $I^t$ and memory
\State Update memory with ($I^t, \hat{M}^t$)
\EndFor
\end{algorithmic}
\label{algorithm1}
\end{algorithm}

\begin{figure*}[t]
\centering
\includegraphics[width=1\linewidth]{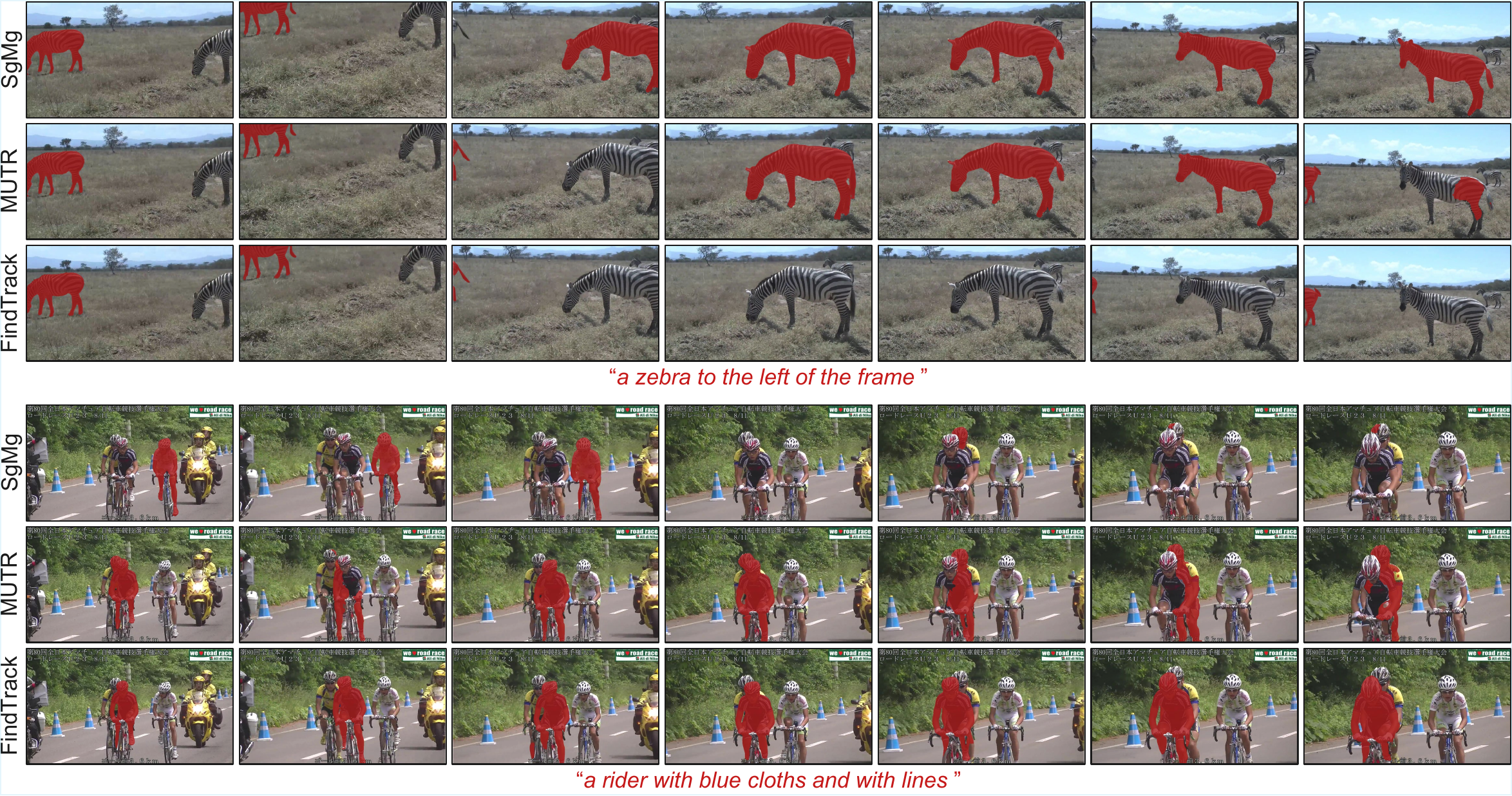}
\vspace{-6mm}
\caption{Qualitative comparison between FindTrack and state-of-the-art methods.}
\label{figure6}
\end{figure*}

\begin{figure*}[t]
\centering
\includegraphics[width=1\linewidth]{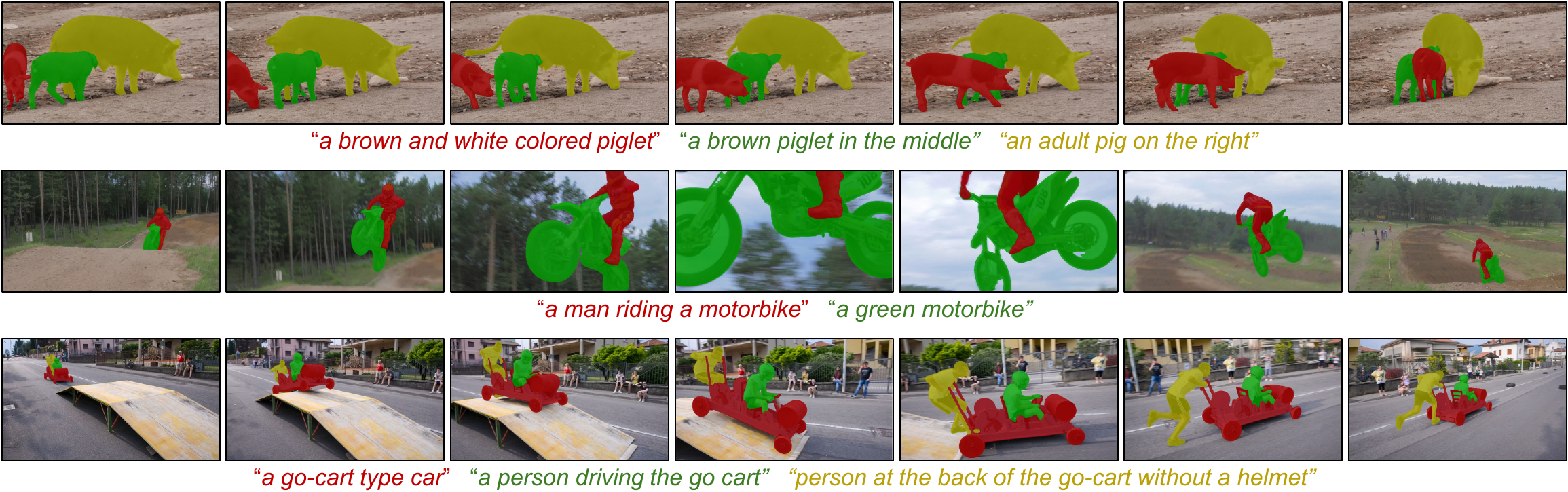}
\vspace{-6mm}
\caption{Qualitative results of FindTrack in challenging scenarios.}
\vspace{-2mm}
\label{figure7}
\end{figure*}

\subsection{Implementation Details}

\noindent\textbf{Model setup.}
FindTrack consists of three core networks: EVF-SAM~\cite{EVF-SAM}, Alpha-CLIP~\cite{AlphaCLIP}, and Cutie~\cite{Cutie}. EVF-SAM, built on a BEiT~\cite{BEiT} backbone, provides language guidance to the SAM~\cite{SAM} decoder and is trained on standard RIS datasets such as RefCOCO~\cite{RC1, RC2, RC3, RC4} and ADE20K~\cite{ADE20K}. Alpha-CLIP is trained on GRIT-20M~\cite{GRIT20m} with RGBA region-text pairs, while Cutie is trained under the MEGA setting, integrating multiple VOS datasets~\cite{DAVIS, YTVOS, MOSE, BURST, OVIS}. Notably, our system can operate without additional training, as the roles of each component are effectively decoupled and seamlessly integrated, fully leveraging the knowledge acquired from each task.

\vspace{1mm}
\noindent\textbf{Inference setup.} 
In the target identification stage, we uniformly sample $N$ frames from the video sequence as candidate key frames for temporal propagation, setting $N$ = 5 by default to balance performance and inference speed. The mask score $\sigma$ for each candidate frame is computed as a weighted sum of the segmentation confidence score $\pi$ and the vision-text alignment score $\rho$, with weights $\mathrm{w}_1$ = 0.5 and $\mathrm{w}_2$ = 0.5. The frame with the highest mask score is selected as the key frame. For temporal propagation, memory in the tracking model is updated every three frames, with long-term memory extraction disabled by default.

We maintain a consistent inference setup across all benchmarks, including candidate frame sampling, key frame selection criteria, and memory update policy. Two exceptions apply: 1) on Ref-YouTube-VOS~\cite{URVOS}, where frames are provided at five-frame intervals, memory is updated accordingly; 2) on MeViS~\cite{MeViS}, which contains extremely long videos, long-term memory is enabled. While this standardized setup ensures generalizability, further tuning may improve performance.

Our model runs on a single GeForce RTX 3090 GPU. Under the representative setting, processing a 30-frame video clip requires approximately 7.9 GB of GPU memory, with an inference speed of around 10 fps.

\section{Experiments}

\subsection{Datasets}
We evaluate FindTrack on three established referring VOS benchmarks: Ref-YouTube-VOS~\cite{URVOS}, Ref-DAVIS17~\cite{RefDAVIS}, and MeViS~\cite{MeViS}. Ref-YouTube-VOS extends YouTube-VOS~\cite{YTVOS} with referring annotations, providing 202 validation videos. Ref-DAVIS17 is built on DAVIS~\cite{DAVIS}, comprising 30 videos, where each object is annotated with four language expressions. MeViS~\cite{MeViS} is a large-scale dataset designed for motion-aware referring segmentation, featuring 140 validation videos and 2,236 language expressions.


\begin{table*}[t]
\centering 
\caption{Quantitative evaluation on the Ref-YouTube-VOS validation set, Ref-DAVIS17 dataset, and MeViS validation set.}
\vspace{-2mm}
\small
\begin{tabular}{p{3.0cm}|P{1.8cm}|P{8mm}P{8mm}P{8mm}|P{8mm}P{8mm}P{8mm}|P{8mm}P{8mm}P{8mm}}
\toprule
\multirow{2}{*}{Method} & \multirow{2}{*}{Publication} & \multicolumn{3}{c|}{Ref-YouTube-VOS} & \multicolumn{3}{c|}{Ref-DAVIS17} & \multicolumn{3}{c}{MeViS} \\
& & $\mathcal{J} \& \mathcal{F}$ & $\mathcal{J}$ & $\mathcal{F}$ 
& $\mathcal{J} \& \mathcal{F}$ & $\mathcal{J}$ & $\mathcal{F}$ 
& $\mathcal{J} \& \mathcal{F}$ & $\mathcal{J}$ & $\mathcal{F}$\\
\midrule
ReferFormer~\cite{ReferFormer} &CVPR'22 &62.9 &61.3 &64.6 &61.1 &58.1 &64.1 &31.0 &29.8 &32.2\\
SgMg~\cite{SgMg} &ICCV'23 &65.7 &63.9 &67.4 &63.3 &60.6 &66.0 &- &- &-\\
HTML~\cite{HTML} &ICCV'23 &63.4 &61.5 &65.2 &62.1 &59.2 &65.1 &- &- &-\\
SOC~\cite{SOC} &NeurIPS'23 &67.3 &65.3 &69.3 &65.8 &62.5 &69.1 &- &- &-\\
VLT+TC~\cite{VLT2} &TPAMI'23 &62.7 &- &- &60.3 &- &- &35.5 &33.6 &37.3\\
MUTR~\cite{MUTR} &AAAI'24 &68.4 &66.4 &70.4 &68.0 &64.8 &71.3 &- &- &-\\
LoSh~\cite{LoSh} &CVPR'24 &67.2 &65.4 &69.0 &64.3 &61.8 &66.8 &- &- &-\\
DsHmp~\cite{DsHmp} &CVPR'24 &67.1 &65.0 &69.1 &64.9 &61.7 &68.1 &46.4 &43.0 &49.8\\
VD-IT~\cite{VD-IT} &ECCV'24 &66.5 &64.4 &68.5 &69.4 &66.2 &72.6 &- &- &-\\
DMVS~\cite{DMVS} &CVPR'25 &64.3 &62.4 &66.2 &65.2 &62.2 &68.2 &48.6 &44.2 &52.9\\
SSA~\cite{SSA} &CVPR'25 &64.3 &62.2 &66.4 &67.3 &64.0 &70.7 &48.6 &44.0 &53.2\\
SAMWISE~\cite{SAMWISE} &CVPR'25 &69.2 &67.8 &70.6 &70.6 &67.4 &74.5 &49.5 &46.6 &52.4\\
\midrule
\textbf{FindTrack} ($N$ = 5) & &70.3 &68.6 &72.0 &\textbf{74.2} &\textbf{69.9} &\textbf{78.5} &47.0 &44.3 &49.7\\
\textbf{FindTrack} ($N$ = 10) & &70.3 &68.6 &71.9 &73.7 &69.4 &78.0 &48.2 &45.6 &50.7\\
\textbf{FindTrack++} ($N$ = 5) & &73.1 &71.2 &75.0 &- &- &- &52.1 &49.4 &54.9\\
\textbf{FindTrack++} ($N$ = 10) & &\textbf{73.7}  &\textbf{71.8} &\textbf{75.7} &- &- &-  &\textbf{53.2} &\textbf{50.5} &\textbf{55.9}\\
\bottomrule
\end{tabular}
\label{table1}
\end{table*}

\subsection{Qualitative Results}
In Figure~\ref{figure6}, we compare FindTrack with state-of-the-art methods, SgMg~\cite{SgMg} and MUTR~\cite{MUTR}, on the Ref-YouTube-VOS~\cite{URVOS} dataset. The first example illustrates a case where the target object exits the frame due to significant camera movement, with visually similar objects appearing in subsequent frames. The second and third examples depict scenarios where the target object becomes entangled with challenging visual distractors. In both cases, SgMg and MUTR struggle to correctly identify the target, as they fail to effectively leverage temporal continuity, making them susceptible to confusion with similar-looking objects. In contrast, FindTrack constructs a robust reference from a global temporal perspective, enabling accurate target identification and consistent mask propagation throughout the video.

Figure~\ref{figure7} presents qualitative results of FindTrack on the Ref-DAVIS17~\cite{RefDAVIS} dataset, demonstrating its robustness in complex scenarios involving multiple entangled objects, large camera movements, and background clutter, which can significantly degrade the performance of existing methods. FindTrack maintains consistent segmentation quality across these challenging conditions, highlighting its resilience to scene complexity.

\subsection{Quantitative Results}
In Table~\ref{table1}, we compare the performance of FindTrack with state-of-the-art methods on the Ref-YouTube-VOS~\cite{URVOS}, Ref-DAVIS17~\cite{RefDAVIS}, and MeViS~\cite{MeViS} datasets. With the default setting of $N = 5$, FindTrack outperforms all existing methods, achieving improvements of 1.9\%, 6.2\%, and 0.6\% on these datasets, respectively. On the MeViS validation set, which contains long video sequences and thus sparse reference searching, increasing $N$ to 10 yields an additional 1.0\% performance gain by increasing the density of candidate frame sampling during the target identification stage.

Although FindTrack achieves competitive performance without additional training, we further fine-tune the per-frame segmentation network on Ref-YouTube-VOS and MeViS, where training sets are available (FindTrack++). This fine-tuning further boosts performance, surpassing existing state-of-the-art methods by a large margin.

\begin{figure*}[t]
\centering
\includegraphics[width=1\linewidth]{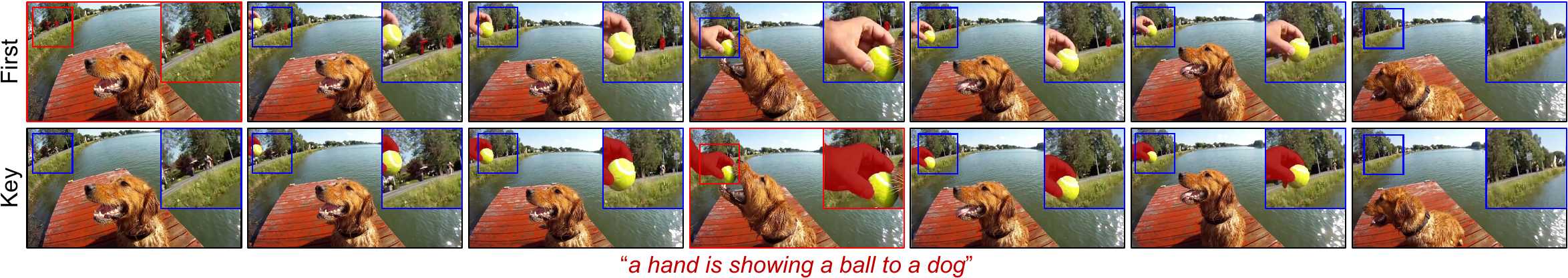}
\vspace{-6mm}
\caption{Visualization of results from different reference selection protocols, with the reference highlighted in \textcolor{red}{red}.}
\label{figure8}
\end{figure*}

\begin{figure*}[t]
\centering
\includegraphics[width=1\linewidth]{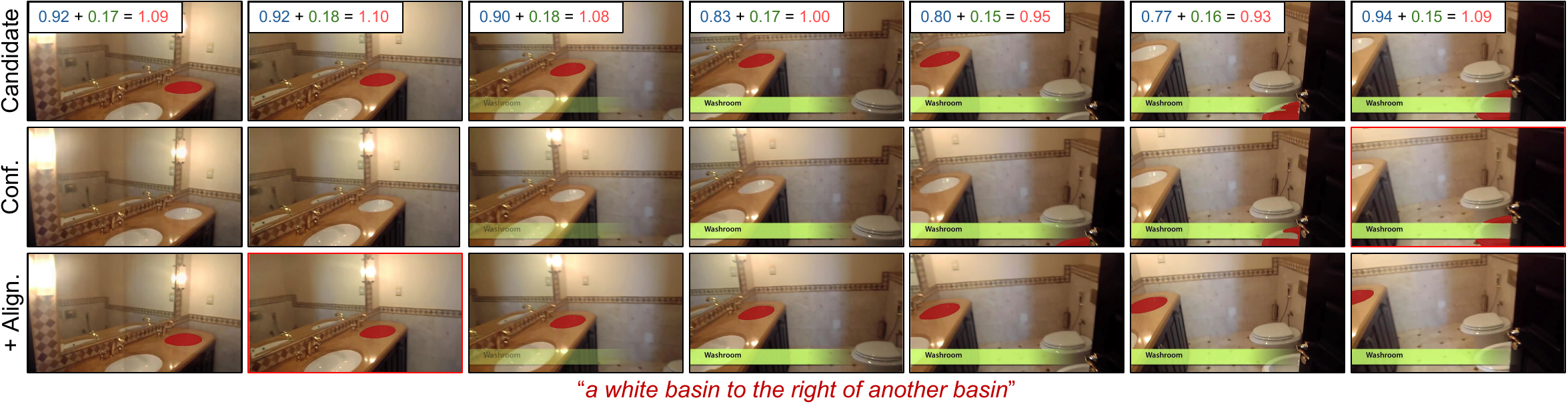}
\vspace{-6mm}
\caption{Visualization of candidate key frames with their corresponding scores, along with the final results using different mask score calculation protocols. The scores are color-coded in accordance with Figure~\ref{figure3}, with the reference frame highlighted in \textcolor{red}{red}.}
\label{figure9}
\end{figure*}

\begin{table}[t]
\centering 
\caption{Ablation study on reference selection.}
\vspace{-2mm}
\small
\begin{tabular}{P{2cm}|P{8mm}P{8mm}|P{8mm}P{8mm}}
\toprule
\multirow{2}{*}{Reference} & \multicolumn{2}{c|}{Ref-YTVOS} & \multicolumn{2}{c}{MeViS} \\
 & $\mathcal{J}$ & $\mathcal{F}$ & $\mathcal{J}$ & $\mathcal{F}$ \\
\midrule
First Frame &63.4 &67.2 &38.7 &44.7\\
Last Frame &59.0 &62.0 &38.7 &44.4\\
\midrule
Key Frame &68.6 &72.0 &50.5 &55.9\\
\bottomrule
\end{tabular}
\label{table2}
\end{table}

\subsection{Analysis}
\noindent\textbf{Reference selection.} 
Defining an effective reference for temporal propagation is crucial, as it underpins target object segmentation throughout the video. To assess the impact of different reference selection strategies, we compare several protocols in Table~\ref{table2} on the Ref-YouTube-VOS~\cite{URVOS} and MeViS~\cite{MeViS} datasets. The first frame protocol simulates direct integration of an RIS network with a semi-supervised VOS network, while the last frame serves as an alternative baseline. The key frame refers to the frame selected by our candidate search strategy. The results demonstrate that selecting a reference from a temporally global perspective enhances the reliability of per-frame predictions, leading to improved video segmentation performance.

Figure~\ref{figure8} provides a visual comparison between using the first frame and our key frame strategy as references. When the target object, specified by the language prompt, does not appear early in the video, using the first frame as a reference fails to provide a reliable cue, often resulting in the segmentation of an incorrect object. Consistent with the quantitative analysis, segmentation quality varies significantly depending on the visual characteristics of the reference, underscoring the importance of effective reference selection.


\begin{table}[t]
\centering 
\caption{Ablation study on the mask score weights.}
\vspace{-2mm}
\small
\begin{tabular}{P{8mm}|P{8mm}|P{8mm}P{8mm}P{8mm}}
\toprule
$\mathrm{w}_1$ &$\mathrm{w}_2$ &$\mathcal{J} \& \mathcal{F}$ & $\mathcal{J}$ & $\mathcal{F}$\\
\midrule
1.0 &0.0 &69.9 &68.2 &71.7\\
0.8 &0.2 &70.0 &68.3 &71.7\\
0.5 &0.5 &70.3 &68.6 &72.0\\
0.2 &0.8 &69.7 &68.0 &71.6\\
0.0 &1.0 &67.3 &65.4 &69.1\\
\bottomrule
\end{tabular}
\vspace{-2mm}
\label{table3}
\end{table}

\vspace{1mm}
\noindent\textbf{Mask score derivation.} 
Reference selection among candidate key frames is guided by two main criteria: segmentation confidence and vision-text alignment scores. The final reference is determined by a weighted sum of these scores. Table~\ref{table3} compares the performance of different mask score weightings on the Ref-YouTube-VOS validation set. The results show that combining both metrics yields better performance than using either score alone, as they effectively complement each other. Empirically, we find that setting $\mathrm{w}_1$ and $\mathrm{w}_2$ to 0.5 provides the best results.

Figure~\ref{figure9} further demonstrates the effectiveness of our mask score derivation protocol. The first row presents candidate key frames with their segmentation confidence and vision-text alignment scores. The second and third rows display the VOS results using the selected key frame as the reference. When only segmentation confidence is used for tracking (second row), incorrectly segmented frames may be chosen as references, leading to error propagation. In contrast, incorporating the vision-text alignment score (third row) helps filter out inaccurate masks by evaluating their consistency with the language prompt, resulting in more reliable reference selection and improved video segmentation quality.

\begin{figure}[t]
\centering
\includegraphics[width=0.95\linewidth]{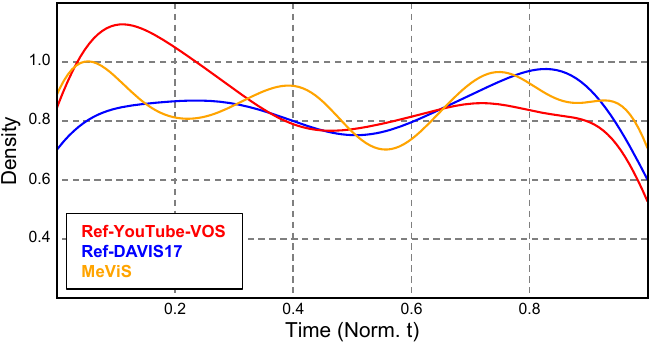}
\vspace{-2mm}
\caption{Visualization of key frame locations.}
\label{figure10}
\vspace{-2mm}
\end{figure}

\vspace{1mm}
\noindent\textbf{Key frame distribution.} 
We visualize the sampled key frame locations in Figure~\ref{figure10} using kernel density estimation. The key frames are quite evenly distributed throughout the video, indicating that the reference selection process is both unbiased and well-balanced.


\section{Conclusion}
We present FindTrack, a groundbreaking approach for referring VOS that decisively decouples target identification from temporal propagation, outperforming all existing methods on public benchmark datasets. By setting a new standard in referring VOS, FindTrack lays a robust foundation for future advancements and innovations in the field.

{
    \small
    \bibliographystyle{ieeenat_fullname}
    \bibliography{FindTrack}
}

\end{document}